\documentclass[conference]{IEEEtran}

\usepackage[utf8]{inputenc}
\usepackage{cite}
\usepackage[linesnumbered,vlined,ruled,boxed,commentsnumbered]{algorithm2e}
\usepackage{algpseudocode}
\usepackage{bm} 

\usepackage{amsmath,amssymb,amsfonts}
\usepackage{commath}
\usepackage{float}

\usepackage[flushleft]{threeparttable}
\usepackage{cleveref}

\usepackage{graphics,graphicx}
\usepackage{subfigure}

\newcommand{\fref}[1]{Fig. \ref{#1}}
\newcommand{\tref}[1]{Table \ref{#1}}

\newcommand{\sref}[1]{Section \ref{#1}}
\newcommand{\eref}[1]{Eq. \ref{#1}}
\newcommand{\aref}[1]{Alg. \ref{#1}}

\usepackage{xcolor} %

\begin{document}

\title{A Semi-Supervised Self-Organizing Map for Clustering and Classification}

\author{
  \IEEEauthorblockN{Pedro H. M. Braga, \textit{Member}, \textit{IEEE}, and Hansenclever F. Bassani, \textit{Member}, \textit{IEEE}}
  \IEEEauthorblockA{Center of Informatics - CIn, Universidade Federal de Pernambuco, Recife, PE, Brazil, 50.740-560\\
  Email: \{phmb4, hfb\}@cin.ufpe.br}
}

\maketitle

\begin{abstract}
There has been an increasing interest in semi-supervised learning in the recent years because of the great number of datasets with a large number of unlabeled data but only a few labeled samples. Semi-supervised learning algorithms can work with both types of data, combining them to obtain better performance for both clustering and classification. Also, these datasets commonly have a high number of dimensions.  This article presents a new semi-supervised method based on self-organizing maps (SOMs) for clustering and classification, called Semi-Supervised Self-Organizing Map (SS-SOM). The method can dynamically switch between supervised and unsupervised learning during the training according to the availability of the class labels for each pattern. Our results show that the SS-SOM outperforms other semi-supervised methods in conditions in which there is a low amount of labeled samples, also achieving good results when all samples are labeled.
\end{abstract}

\IEEEpeerreviewmaketitle

\section{Introduction}


In recent years, research on Artificial Neural Networks with supervised learning algorithms has made great advances, often appearing in technology news with increasingly impressive practical applications in diverse areas, such as Robotics \cite{Levine2016}, Genomics \cite{araujo2013learning}, and Natural Language Processing \cite{Zhou2016}.

Despite these advances, the fact that these methods require a large amount of properly labeled data for training (sometimes, in the order of thousands of patterns per class) makes their use in many applications impractical. In certain areas, such as in the medical field, it is extremely difficult and expensive to obtain balanced labeled datasets.  In other areas, such as robotics, the dynamic imposed makes it impossible to have real-time labels. In addition, in certain problems, new categories of elements may frequently arise, making it infeasible to create a comprehensive previously labeled training dataset.
 
Therefore, at the current stage of research, it is of great importance to put forward methods that can benefit both from the (frequently large amounts of) unlabeled data available as well as from the smaller amounts of labeled data, what would expand the current range of machine learning applications.

In order to achieve performance improvements, Semi-Supervised Learning (SSL) methods take advantage of both unlabeled and labeled data \cite{label-propagation}. Hence, SSL is halfway between supervised and unsupervised learning, being applied for both classification and clustering tasks \cite{basu2002semi}.

In semi-supervised classification, the training process tries to exploit additional information (often available as label classes) together with the unlabeled data to achieve a more accurate classification function. In semi-supervised clustering, this prior information is used to obtain a better clustering performance \cite{basu2002semi,schwenker2014pattern}.
Prototype-based methods such as K-Means \cite{basu2002semi} and Self-Organizing Maps (SOM) \cite{kohonen1990,bassani2015larfdssom} are examples that have been successfully applied in this area. 

Kohonen proposed two very influential prototype-based methods. SOM \cite{kohonen1990} is an unsupervised learning method frequently applied for clustering, and the Learning Vector Quantization (LVQ) \cite{kohonen1995learning} is a supervised learning method that shares many similarities with SOM, which is frequently applied for classification. Therefore, these methods are good candidates for developing a hybrid approach for SSL. 

Various modifications of LVQ and SOM were proposed to improve their performance in more challenging datasets with thousands of dimensions, commonly found in areas such as data mining \cite{kriegel2009clustering} and bioinformatics \cite{araujo2013learning}. In this context, traditional distance metrics often applied in prototype-based methods may become meaningless due to the curse of dimensionality \cite{koppen2000curse}, in which objects may appear approximately equidistant from each other, what is aggravated by the presence of irrelevant dimensions in the dataset. SOM and LVQ-based methods usually deal with such problems by applying weights to the input dimensions, what has been shown to provide significant performance improvements. 

Following this path, in this paper, we proposed a new method called Semi-Supervised Self-Organizing Map (SS-SOM), which is an extension of Local Adaptive Receptive Field Dimension Selective Self-Organizing Map (LARFDSSOM) \cite{bassani2015larfdssom}, created by introducing important modifications to incorporate semi-supervised learning. 

In order to evaluate the SS-SOM, we compared it with other supervised and semi-supervised methods. The performance of SS-SOM was evaluated in different conditions of labels availability, ranging from 1\% to 100\% of labeled samples in the dataset. The proposed method presents promising results when applied to real-world datasets, even in conditions of a low percentage of labeled data, reaching a similar accuracy of traditional supervised learning methods.

The rest of this article is structured as follows: \sref{machine-learning} defines the machine learning approaches considered in this article. \sref{related-work} presents a review of important and prominent classification and clustering methods from different learning approaches. \sref{prop-method} describes in details the proposed method. \sref{experiments} presents the experimental setup, methodology and the obtained results and comparisons. Finally, in \sref{conclusions} we discuss the obtained results and indicate future directions.

\section{Machine Learning Approaches}
\label{machine-learning}
In a broad sense, the learning processes are traditionally categorized into two fundamentally different types of tasks: learning with and without supervisor \cite{mlp,chapelle2009semi}.

In the first, called supervised learning, involving only labeled data, the goal is to learn a mapping from \textit{X} to \textit{Y}, given a training set made of pairs ($x_i$, $y_i$), where $y_i \in Y$ are the labels of the samples $x_i$. The latter, involving only unlabeled data, can be divided into two subcategories: 1) unsupervised learning, where the goal is to find interesting structure in the data X by estimating a density of which is likely to have generated X; and 2) reinforcement learning, where the learning of an input-output mapping is performed through continued interaction with the environment in order to minimize some kind of cost function \cite{mlp,chapelle2009semi}.

In the past years, there is a growing interest in a hybrid setting, called semi-supervised learning (SSL). SSL is a central point between supervised and unsupervised learning. In many learning tasks, there is a large supply of unlabeled data, but insufficient labeled ones, since it can be expensive and hard to generate. The basic idea of SSL is to take advantage of both labeled and unlabeled data during the training, combining them to improve the performance of the models \cite{schwenker2014pattern,jain2010data,chapelle2009semi,basu2002semi}.

Moreover, SSL can be further classified into semi-supervised classification and semi-supervised clustering \cite{schwenker2014pattern}. Firstly, in the semi-supervised classification, the training set is given in two parts: $S = \{(\textbf{x}_i, y_i) | \textbf{x}_i \in \mathbb{R}^d, y_i \in Y, 1 \leq i \leq M \}$ and $U = \{ \textbf{u}_i \in  \mathbb{R}^d | i = 1, \cdots, M \}$. Where \textit{S} and \textit{U} are the labeled and unlabeled data, respectively. At first hand, it is possible to consider a traditional supervised scenario using only \textit{S} to build a classifier. However, the unsupervised estimation of the probability function \textit{p}(\textbf{x}) of the input set can take advantage of both \textit{S} and \textit{U}. Besides, classification tasks can reach a higher performance through the use SSL as a combination of supervised and unsupervised learning \cite{schwenker2014pattern}. Many semi-supervised classification algorithms have been developed in the past decades, and, according to Zhu \cite{zhu2006semi}, we can structure them into the following categories: 1) Self-training; 2) SSL with generative models; 3) Semi-supervised Support Vector Machines ($S^3$VM), or transductive SVM; 4) SSL with graphs; and 5) SSL with committees.

Secondly, in the semi-supervised clustering, the aim is to group the data in an unknown number of groups relying on some kind of similarity or distance measures in combination with objective functions. Clustering is a more difficult and challenging problem than classification, and the nature of the data can make the clustering tasks even more difficult, so any kind of additional prior information in respect to the data can be useful to obtain a better performance. Therefore, the general idea behind semi-supervised clustering is to integrate some type of prior information in the process. For example, a subset of labeled data and further constraints on pairs of the patterns in form of \textit{must-link} and \textit{cannot-link} \cite{schwenker2014pattern,zhu2006semi}.  Prototype-based models algorithms (e.g., k-means, and SOMs), Hidden Markov Random Fields (HMRFs), Expectation Maximization (EM) and Label Propagation (LP) are examples that have been successful in this area \cite{schwenker2014pattern,zhu2006semi,basu2002semi,jain2010data}.
\section{Related Work}
\label{related-work}
Several techniques have been developed and used to deal with high-dimensional data in different learning contexts. Thus, in this section, we describe the unsupervised (\sref{unsup-met}), supervised (\sref{sup-met}), and semi-supervised (\sref{ss-met}) methods and discuss how they are connected with the motivating problem. Some of these methods will be further compared in the \Cref{experiments,conclusions}.

\subsection{Unsupervised Methods}
\label{unsup-met}
Unsupervised learning techniques can address the problem imposed by the high-dimensional and unlabeled data. In this context, we can cite the concept of Self-Organizing Maps (SOM), first introduced by Kohonen \cite{kohonen1995learning}. SOM is used in several applications including clustering data without the knowledge of the labels. SOM also provides a topology that preserves the mapping from the high-dimensional space to map units and the relation between the points.

The general task of clustering involves not only clustering the data but also identifying subsets of the input dimensions which are relevant to characterize each cluster. One way to achieve this is by applying local relevances to the input dimensions. The identification of which dimension is relevant or not is an important feature when working with high-dimensional data \cite{araujo2013learning}. In this context, subspace clustering methods have been proposed aiming to determine clusters in subspaces of the input dimensions of a given dataset \cite{kriegel2009clustering}. Moreover, in subspace clustering problems, a sample may belong to more than one cluster as a result of taking into account different subsets of the input dimensions \cite{bassani2015larfdssom}. On the other hand, it is important to mention that in projected clustering problems, each sample belongs to a single cluster. 

Therefore, some variations of the original SOM were developed to improve the performance of the clustering tasks, and LARFDSSOM is an example. It uses a time-varying structure, a neighborhood defined by connecting nodes that have similar subspaces of the input dimensions, and a local receptive field that is adjusted for each node as a function of its local variance. Hence, LARFDSSOM showed good results in the motivating problem for both subspace and projected clustering \cite{bassani2015larfdssom}.

\subsection{Supervised Methods}
\label{sup-met}
Some supervised methods for classification were proposed to deal with high-dimensional data. According to Hammer \cite{hammer2004relevance}, some Learning Vector Quantization (LVQ) methods are good options, since they have been shown to be a valuable alternative to Support Vector Machines (SVMs) \cite{cortes1995support}. Even so, SVMs and the Multilayer Perceptron (MLP) \cite{mlp} are also alternatives.

As the SOM, LVQ was also proposed by Kohonen \cite{kohonen1995learning}. It is a family of algorithms for statistical pattern classification, which uses prototypes to represent class regions \cite{nova2014review}. These regions are defined by hyperplanes between prototypes, resulting in Voronoi partitions. Various modifications of LVQ exist to ensure faster convergence, a better adaptation of the receptive fields, and an adaptation for complex data structures \cite{hammer2002generalized}.

The Generalized Learning Vector Quantization (GRLVQ) is a member of this family. The algorithm was inspired by GLVQ and proposed to deal with high dimensional datasets by using a relevance vector able to identify the irrelevant dimensions and/or noise commonly present in real datasets. GRLVQ adapts weights for each input dimension by incorporating an update rule \cite{hammer2002generalized}.

\subsection{Semi-supervised Methods}
\label{ss-met}
The K-means is one of the most popular and simple clustering algorithms. Despite the fact that K-means was proposed over 50 years ago, it is still widely used, and many variations have been proposed. Semi-supervised K-means-based methods were very successful demonstrating their advantages over standard approaches. One of them is called Seeded-KMeans \cite{basu2002semi}. It can be viewed as an instance of the EM algorithm, where labeled data provides prior information about the conditional distribution of hidden category labels working as a guide for the clustering process.

Given a dataset \textit{X}, the K-means clustering of the dataset generates a number of k partitions of \textit{X}. Let $S \subseteq X$ be the \textit{seed set}, a subset of data-points on which supervision is provided as follows: for each $x_i \in S$, a group $X_i$ will be created according to the partition to with it belongs. By the end of the process, the partitions of the seed set S will form the seed clustering and will be used to guide the K-means algorithm \cite{basu2002semi}.

In the Seeded-KMeans, the seed clustering is used only to initialize the K-means algorithm. Hence, instead of initializing from K random means, the mean of the \textit{i}th cluster is initialized with the mean of the \textit{i}th partition $S_i$ of the seed set. 

Label propagation (LP) is another promising approach for SSL \cite{herrmann2007label}. LP methods operate on proximity graphs or connected structures to spread and propagate information about the class to nearby nodes according to a similarity matrix. It is based on the assumption that nearby entities should belong to the same class, in contrast to far away entities \cite{label-propagation,herrmann2007label}.

For LP purposes, each node is assigned to a label vector. A label vector $l_i \int [0, 1]^k$ contains the probabilistic membership degrees of input samples to the available cluster. Here, the nodes propagate their label vectors to all adjacent nodes according to a defined distance W. Nodes belonging to a pre-classified input sample have fixed label vectors \cite{herrmann2007label}.

A similar alternative to LP is called Label Spreading (LS) \cite{label-spreading}. It differs from LP in modifications to the similarity matrix. LP uses the raw similarity matrix constructed from the data with no changes, whereas LS minimizes a loss function that has regularization properties allowing it to be often better regarding robustness to noise.

\section{Proposed Method}
\label{prop-method}

SS-SOM\footnote{Available at: https://github.com/phbraga/SS-SOM} is a semi-supervised hybrid SOM, based on LARFDSSOM \cite{bassani2015larfdssom}, with a time-varying structure \cite{araujo2013self} and two different ways of learning. It is possible for SS-SOM, as in LARFDSSOM, that the nodes consider different relevances for the input dimensions and adapts its receptive field during the self-organization process. 

Moreover, our model is a prototype-based method that can learn in a supervised or unsupervised way. The SS-SOM can switch between these two ways during the self-organization process according to the availability of the information about the class label for each input pattern. To achieve this, we modified the LARFDSSOM to include concepts from the standard LVQ \cite{kohonen1995learning} when the class label of some input pattern is given. The operations of the map consist of three phases: 1) organization (\aref{alg:hybrid-mode}); 2) convergence; and 3) clustering or classification. 

\begin{algorithm}[!ht]
\small
Initialize parameters $a_{t}$, \textit{lp}, $\beta$, \textit{age\_wins}, $e_{b}$, $e_{n}$, $\epsilon\beta$, \textit{minwd}, $t_{max}$, \textit{push\_rate},  $N_{max}$;

Initialize the map with one node with $\textbf{c}_{j}$ initialized at the first input pattern $\textbf{x}_{0}$, $\boldsymbol{\omega_j}$ $\gets$ \textbf{1}, $\boldsymbol{\delta_j}$ $\gets$ \textbf{0}, $\text{wins}_j$ $\gets$ 0 and $\text{class}_j$ $\gets$ \textit{noClass} or \textit{class}($\textbf{x}_{0}$) if available;

Initialize the variable nwins $\gets$ 1;

\For{t $\gets$ 0 \textit{\textbf{to}} $t_{max}$}
{
	Choose a random input pattern \textbf{x};

	Compute the activation of all nodes (\eref{activation_dssom});

	Find the winner $s_1$ with the highest activation ($a_s$) (\eref{winner_dssom});
    
	\eIf{\text{\textbf{x} has a label}}
    { 
    	Run the SupervisedMode(\textbf{x}, $s_1$) (\aref{alg:supervised-mode});
        
    } {
    	Run the UnsupervisedMode(\textbf{x}, $s_1$) (\aref{alg:unsupervised-mode});
    }
    
    \If{\textit{nwins} = \textit{age\_wins}}
    {
    	Remove nodes with $\text{wins}_{j}$ $<$ \textit{lp} $\times$ \textit{age\_wins};
        
    	Update the connections of the remaining nodes (\eref{update_neighborhood_larfdssom});
        
        Reset the number of wins of the remaining nodes:
        
        $\text{wins}_{j}$ $\gets$ 0;
        
        \textit{nwins} $\gets$ 0;
    }
    
    \textit{nwins} $\gets$ \textit{nwins} + 1;
}

Run the Convergence Phase;
\caption{Hybrid Mode}
\label{alg:hybrid-mode}
\end{algorithm}

In the organization phase, after the network initialization, the nodes compete to form clusters of randomly chosen input patterns. There are two different ways to decide who is the winner of a competition, which nodes need to be updated and when a new node needs to be inserted. If the class label of the input pattern is provided, the supervised mode is used (\sref{sup-mode}), otherwise, the unsupervised mode is employed (\sref{unsup-mode}). The model can be trivially modified to also incorporating reinforcement learning. The neighborhood of SS-SOM is formed connecting nodes with others of the same class label, or with unlabeled nodes. In both cases, it is necessary to take into account a similar subset of the input dimensions. The competition, adaptation and cooperation steps are repeated for a limited number of epochs. Furthermore, as in LARFDSSOM, the nodes that do not win for a minimum number of patterns are removed from the map every time that a certain age number ($age\_wins$ parameter) is reached.

The convergence phase starts after the organization phase. Here, the nodes are also updated and removed when necessary, similarly to the way conducted in the first phase. The difference is the fact that there is no insertion of new nodes. Moreover, this phase finishes the cycle left by the organization phase and runs another one to ensure convergence.

After finishing the convergence phase, the map can cluster and classify input patterns. Depending on the amount and distribution of labeled input patterns presented to the network during the training, after the convergence phase the map may have: 1) all the nodes labeled; 2) some nodes labeled; 3) no nodes labeled. For the first case, the clustering and classification are straightforward: each test pattern is associated with the label of the node with the highest activation. For the second case, if the node with the highest activation has no class, we continue looking for another node with a defined class label, and an activation above the threshold $a_t$. For the third and final case, we can identify the clusters of the input patterns, but not their classes.

It is important to mention that in subspace clustering an input pattern may belong to more than one cluster. However, in this work, we considered only the task of projected clustering, when each input pattern is assigned to a single cluster.

The next sections describe the operation in the unsupervised and supervised modes.

\subsection{Unsupervised Mode}
\label{unsup-mode}
Given an unlabeled input pattern, we look for a winner node disregarding their class labels. Therefore, as in the \eref{winner_dssom}, the winner of a competition is the node that is the most activated according to a radial basis function with the receptive field adjusted as a function of its relevance vector. In other words, the winner \textit{s}(\textbf{x)} is the node with the highest activation value (\sref{nodes-at}) for the input pattern:

\begin{equation}
\small
\label{winner_dssom}
s(\textbf{x}) = \text{arg} \max_j [ac(D_{\omega}(\textbf{x}, \textbf{c}_j), \omega_j)],
\end{equation}
where $ac(D_{\omega}(\textbf{x}, \textbf{c}_j)$ is the activation function explained in \sref{nodes-at} and $\omega_j$ is the relevance vector of the node \textit{j}.

Similarly to LARFDSSOM, SS-SOM has an activation threshold $a_t$. According to this, if the activation of the winner is lower than $a_t$, a new node is inserted into the map at the input pattern position because the winner is not close enough. Otherwise, the winner and its neighbors are updated to get closer to the input pattern (\sref{node-update}), for that, we consider two fixed learning rates: 1) $e_b$ for the winner node; and 2) $e_n$ for its neighbors, where $e_n$<$e_b$. \aref{alg:unsupervised-mode} presents this procedure.

\begin{algorithm}[!ht]
\small

\SetKwInOut{Input}{Input}
\Input{Input pattern \textbf{x} and the first winner $s_1$;}
\eIf{$a_{s_1}$ $<$ $a_{t}$ and \textit{N} $<$ $\textit{N}_{max}$}
{
	Create new node \textit{j} and set: $\textbf{c}_{j}$ $\gets$ \textbf{x}, $\boldsymbol{\omega_j}$ $\gets$ \textbf{1}, $\boldsymbol{\delta_j}$ $\gets$ \textbf{0}, $\text{wins}_j$ $\gets$ 0 and $\text{class}_j$ $\gets$ \textit{noClass};

	Connect j to the other nodes as per \eref{update_neighborhood_larfdssom};
}{ 

	Update the winner node and its neighbors: UpdateNode($s_1$, $e_b$), UpdateNode(\textit{neighbors}($s_1$), $e_n$) (\aref{alg:update-node});
    
    Set $\text{wins}_{s_1}$ $\gets$ $\text{wins}_{s_1}$ + 1;
}
\caption{Unsupervised Mode}
\label{alg:unsupervised-mode}
\end{algorithm}

\subsection{Supervised Mode}
\label{sup-mode}
In order to incorporate the supervised learning mode, each node in the map can be associated with a class label. Hence, when a labeled input pattern is given, we treat it differently. The \aref{alg:supervised-mode} presents this procedure.

\begin{algorithm}[!ht]
\small

\SetKwInOut{Input}{Input}
\Input{Input pattern \textbf{x} and the first winner $s_1$;}

\eIf{$\text{class}_{s_1}$ = \textit{class}(\textbf{x}) \textbf{or} $\text{class}_{s_1}$ = noClass}
{
	\If{$a_{s_1}$ $<$ $a_{t}$ and \textit{N} $<$ $\textit{N}_{max}$}
	{
    	Create new node \textit{j} and set: $\textbf{c}_{j}$ $\gets$ \textbf{x}, $\boldsymbol{\omega_j}$ $\gets$ \textbf{1}, $\boldsymbol{\delta_{j}}$ $\gets$ \textbf{0}, $\text{wins}_j$ $\gets$ 0 and $\text{class}_{j}$ $\gets$ \textit{class}(\textbf{x});
        
        Connect j to the other nodes as per \eref{update_neighborhood_larfdssom};
	}
    \ElseIf{$a_{s_1}$ $\geq$ $a_{t}$}
    { 
    	Update the winner node and its neighbors: UpdateNode($s_1$, $e_b$), UpdateNode(\textit{neighbors}($s_1$), $e_n$) (\aref{alg:update-node});
        
    	Set $\text{class}_{s_1}$ $\gets$ \textit{class}(\textbf{x});
        
        Update $s_1$ connections as per \eref{update_neighborhood_larfdssom};
        
    	Set $\text{wins}_{s_1}$ $\gets$ $\text{wins}_{s_1}$ + 1;
    }
}{ 
	Try to find a new winner $s_2$ with \textit{noClass} or the same class of \textbf{x} with activation $a_{s_2}$ $\geq$ $a_{t}$;
    
    \If{$s_2$ exists}
    {
    
    	Update the new winner node, its neighbors and the previous wrong winner: UpdateNode($s_2$, $e_b$), UpdateNode(\textit{neighbors}($s_2$), $e_n$) and UpdateNode($s_1$, -push\_rate) (\aref{alg:update-node});
        
        Set $\text{wins}_{s_2}$ $\gets$ $\text{wins}_{s_2}$ + 1;
    }
    \ElseIf{\textit{N} $<$ $\textit{N}_{max}$}
    {
    	Create new node \textit{j} and set: $\textbf{c}_{j}$ $\gets$ \textbf{x}, $\boldsymbol{\omega_j}$ $\gets$ \textbf{1}, $\boldsymbol{\delta_{j}}$ $\gets$ \textbf{0}, $\text{wins}_j$ $\gets$ 0 and $\text{class}_{j}$ $\gets$ \textit{class}(\textbf{x});
        
        Connect $j$ to other nodes as per \eref{update_neighborhood_larfdssom};
    }
	
}

\caption{Supervised Mode}
\label{alg:supervised-mode}
\end{algorithm}

In order to obtain performance improvements from the labeled patterns, we take the labels into account when looking for a winner. Here, unlike the unsupervised mode that only consider the activation, if the most activated node $\textit{s}_1$ has the same class of the input pattern or a not defined class (line 1 in \aref{alg:supervised-mode}), a very similar procedure to the unsupervised mode (\sref{unsup-mode}) is ran (lines 2 to 9). The difference, in this case, is the fact that is necessary to set $\textit{s}_1$ class to the same class of the input pattern \textbf{x}, as well as update its connections. Otherwise, we search for another winner matching the following conditions (line 11): 1) it needs to have the same class of the input pattern or an unspecified class, and 2) the activation must be higher than $a_t$.

If any node fulfills these conditions (line 12  in \aref{alg:supervised-mode}), a new winner $\textit{s}_2$ has been found, and it and its neighbors will be updated as in the unsupervised mode (\sref{unsup-mode}). However, the fact that $\textit{s}_1$ was the wrong winner leads to the possibility to push it away from the input pattern. Therefore, similarly as in the LVQ, we push $\textit{s}_1$ away from the input pattern with a fixed learning rate $-push\_rate$. This procedure is presented in lines 13 and 14 of \aref{alg:supervised-mode}. Otherwise, if the maximum number of nodes in the map was not achieved, a new node is inserted into the map at the same position and with the same class of the input pattern \textbf{x} (lines 16 and 17 of \aref{alg:supervised-mode}).

\subsection{Common Operations for Both Modes}
\subsubsection{Nodes structure}
In SS-SOM, each node \textit{j} in the map represents a cluster and is associated with three \textit{m}-dimensional vectors, where \textit{m} is the number of input dimensions; $ \textbf{c}_j = \{c_{ji}, i = 1, \cdot\cdot\cdot, m\}$ is the center vector that represents the prototype of the cluster \textit{j} in the input space; $\boldsymbol\omega_j = \{\omega_{ji}, i = 1, \cdot\cdot\cdot, m\}$ is the relevance vector in which each component represents the estimated relevance, a weighting factor within [0, 1], that the node \textit{j} applies for the \textit{i}th input dimension; and $\boldsymbol\delta_j = \{\delta_{ji}, i = 1, \cdot\cdot\cdot, m\}$ is the distance vector, that stores a moving average of the observed distance between the input patterns \textbf{x} and the center vector $| \textbf{x} - \textbf{c}_j(n)|$. The $\boldsymbol\delta$ vector is used solely to compute the relevance vector, as in \cite{bassani2015larfdssom}.

\subsubsection{Nodes activation}
\label{nodes-at}
The activation of a node in SS-SOM is calculated as a radial basis function of the weighted distance $D_\omega(\textbf{x}, \textbf{c}_j)$ with the receptive field adjusted as a function of its relevance vector. The activation grows as the distance decreases and as the relevances increases. The \eref{activation_dssom} shows the activation calculation as follows:

\begin{equation}
\small
\label{activation_dssom}
ac(D_{\omega}(\textbf{x}, \textbf{c}_j), \omega_j) =  \frac{\sum\limits_{i=1}^m \omega_{ji}}{\sum\limits_{i=1}^m \omega_{ji} + D_{\omega}(\textbf{x}, \textbf{c}_j) + \epsilon},
\end{equation}
where $\epsilon$ is a small value added to avoid division by zero and $D_\omega(\textbf{x}, \textbf{c}_j)$ is the weighted distance function used in LARFDSSOM:

\begin{equation}
\small
\label{distance_orig}
D_{\omega}(\textbf{x}, \textbf{c}_j) =  \sqrt{\sum_{i = 1}^{m} \omega_{ji} {(x_{i} - c_{ji})}^{2}}.
\end{equation}

\subsubsection{Node Update}
\label{node-update}
In SS-SOM, in order to update the vectors associated with the nodes (the winner, the neighbors or the winner of a wrong class), a fixed learning rate is used, depending on the undergoing procedure (\aref{alg:supervised-mode} or \aref{alg:unsupervised-mode}).

\begin{algorithm}[!ht]
\small

\SetKwInOut{Input}{Input}
\Input{Node \textit{s}, Learning Rate \textit{lr}}
 
\SetKwFunction{FUpdate}{UpdateNode}
\SetKwProg{Function}{Function}{:}{}
  
\Function{\FUpdate{\textit{s}, \textit{lr}}}
{
	Update the distance vectors $\boldsymbol{\delta_s}$ of \textit{s} according $lr$ (\eref{dist_vec_larfdssom});
    
	Update the relevance vectors $\boldsymbol{\omega_s}$ of \textit{s} (\eref{relevance_vec_larfdssom});
        
	Update the weight vectors $\textbf{c}_{s}$ of \textit{s} (\eref{weight_vec_larfdssom});
}

\caption{Node Update}
\label{alg:update-node}
\end{algorithm}

\aref{alg:update-node} shows how the update occurs in SS-SOM. Given a learning rate, the node will be updated as in LARFDSSOM. We present the equations as follows:

\begin{equation}
\small
\label{weight_vec_larfdssom}
\textbf{c}_j(n + 1) = \textbf{c}_j(n) + e(\textbf{x} - \textbf{c}_j(n)),
\end{equation}
where \textit{e} is the learning rate.

To compute the relevance vectors, we estimate the average distance of each node to the input pattern that it clusters. As in LARFDSSOM, the distance vectors are updated through a moving average of the observed distance between the input pattern and the current center vector 

\begin{equation}
\small
\label{dist_vec_larfdssom}
\boldsymbol{\delta}_j(n + 1) = (1 - e\beta)\boldsymbol{\delta}_j(n) + e\beta(|\textbf{x} - \textbf{c}_j(n)|),
\end{equation}
where \textit{e} is the learning rate, $\beta$ $\in$ ]0,1[ controls the rate of change of the moving average, and the operator $|$ $\cdot$ $|$ denotes the absolute value, not the norm \cite{bassani2015larfdssom}.  

After updating the distance vector, each component $\omega_{ji}$ of the relevance vector is calculated by an inverse logistic function of the distances $\delta_{ji}$ as follows in \eref{relevance_vec_larfdssom}

\begin{equation}
\small
\label{relevance_vec_larfdssom}
\omega_{ji} = \begin{cases}
  \frac{1}{1 \text{ + exp} \Big( \frac{\delta_{ji\text{mean}} - \delta_{ji}}{s(\delta_{ji\text{max}} - \delta_{ji\text{min}})} \Big)} & \text{if } \delta_{ji\text{min}} \neq \delta_{ji\text{max}} \\
  1 & \text{otherwise},
\end{cases}
\end{equation}
where $\delta_{ji\text{max}}$, $\delta_{ji\text{min}}$, $\delta_{ji\text{mean}}$ are the maximum, the minimum, and the mean of the components of the distance vector $\boldsymbol\delta_j$, respectively. The parameter \textit{s} $>$ 0 controls the slope of the logistic function \cite{bassani2015larfdssom}.

\subsubsection{Node Removal}
In SS-SOM, each node \textit{j} in the map stores a variable $wins_{j}$ that represents the number of the node victories since the last reset. Whenever \textit{age\_wins} is reached, a reset occurs (lines 13-19 in \aref{alg:hybrid-mode}), it means that any nodes which do not win at least the minimum percentage of the competitions \textit{lp} $\times$ \textit{age\_wins} will be removed. After the reset, the number of victories of the remaining nodes is set to zero.  

\subsubsection{Neighborhood Update}
When a reset occurs, and the nodes have been removed, the connections between the remaining nodes must be updated. In SS-SOM, the neighborhood is formed by nodes with the same class or unlabeled nodes that apply similar relevances for the input dimensions, so that, a connection between two nodes means that they cluster patterns with the same class or at least in similar subspaces. \eref{update_neighborhood_larfdssom} considers these similarities between the relevances of every pair of nodes to control this behavior.

\begin{equation}
\small
\label{update_neighborhood_larfdssom}
\text{nodes } i \text{ and } j \text{ are } \begin{cases}
  \text{connected}, & \text{if } \text{( }class(i) = class(j) \text{ or } \\
                    & class(i) = noClass \text{ or } \\
                    & class(j) = noClass \text{ )}\\
                    & \text{ and } \norm{\omega_i - \omega_j} < e\sqrt{m}\\
  \text{disconnected}, & \text{otherwise}
\end{cases}
\end{equation}

\subsection{SS-SOM Parameters Summary}
SS-SOM inherits all parameters from LARFDSSOM and includes a new parameter called $push\_rate$. This parameter provides a specific learning rate for the update of wrong winners as described in \sref{sup-mode}. It means that we have 11 parameters to set up. Despite this being a high number of parameters, a sensitivity analysis showed in \cite{bassani2015larfdssom} revealed that only $a_t$ and \textit{lp} present a high impact on the results. SS-SOM kept this characteristic with the addition of $e_b$ as a new sensitive parameter. So that, we can keep the other parameters values fixed inside the ranges defined in \tref{tab:sshsom-params}, given their marginal influences, including the number of epochs. The parameter $a_t$, however, is crucial. Since it defines the receptive field of nodes, during the training, it affects the number of nodes inserted in the map, as well as the number of patterns regarded as outliers during the clustering and classification phase. The parameter \textit{lp} defines the minimum percentage of input patterns that a node has to cluster for not being removed from the map. This parameter is dataset dependent and has a substantial impact on the results. Finally, the parameter $e_b$ is the learning rate of the winner node, it defines the update step, which depends on the dataset. After a well adjust of $a_t$ and \textit{lp}, it starts to impact the results, but it is not so significant than the other two. A short description for the other parameters can be found in \cite{bassani2015larfdssom}.

\section{Experiments}
\label{experiments}

In order to evaluate the classification capabilities of SS-SOM, we compare it with some traditional supervised methods such as MLP \cite{mlp}, SVM \cite{cortes1995support}, and GRLVQ \cite{hammer2002generalized}. We also compared SS-SOM with the following semi-supervised methods: Label Spreading \cite{label-spreading} and Label Propagation \cite{label-propagation}. Finally, we used seven real-world datasets of the OpenSubspace framework \cite{muller2009evaluating}. It provides real-world datasets adapted from the UCI machine learning repository \cite{asuncion2007uci} as well as an extensive amount of synthetic datasets. A detailed description of the datasets can be found in \cite{muller2009evaluating}.

In \sref{exp-setup}, we present the methodology and the experimental setup, next in \sref{exp-results}, we present the results and analysis necessary to clarify the final conclusions.

\subsection{Experimental Setup}
\label{exp-setup}

For all the algorithms, on each dataset, we used 3-times 3-fold cross-validation. Each method was trained and tested 500 times for each fold with different parameter values sampled from the parameter ranges presented in \Crefrange{tab:grlvq-params}{tab:sshsom-params}, according to a Latin Hypercube Sampling (LHS) \cite{helton2005comparison}, while the best accuracy achieved by each method in each fold was recorded for each dataset. This comprises a total of 752,000 experiments. After that, we calculate the mean and the standard deviation of the best results for each dataset separately. The LHS guarantees the full coverage of the range of each parameter. For our case, the range of each parameter is divided into 500 intervals of equal probability which leads to a random selection of a single value from each interval \cite{bassani2015larfdssom}.

For studying the effects of the different levels of supervision, i. e., the percentage of labeled data, the semi-supervised methods were trained with the following percentages: 1\%, 5\%,  10\%, 25\%, 50\%, 75\%\, and 100\%. The ranges of the parameter for the supervised methods are shown in \Crefrange{tab:grlvq-params}{tab:mlp-params} and the parameter ranges for both semi-supervised methods can be seen in \tref{tab:spread-prop-params}. Finally, the ranges for SS-SOM are shown in \tref{tab:sshsom-params}. The maximum number of nodes for SS-SOM ($\textit{N}_{max}$) was set to be the size of the training set. A detailed description of the parameters of the comparable methods can be found in \cite{hammer2002generalized}, \cite{mlp}, \cite{cortes1995support}, \cite{label-propagation}, and \cite{label-spreading}.

\begin{table}[ht!]
\small
\renewcommand{\arraystretch}{1.3}
\caption{Parameter Ranges for GRLVQ}
\label{tab:grlvq-params}
\centering
\begin{tabular}{lcc}
\hline
\bfseries  Parameters & \bfseries min & \bfseries max\\
\hline
Number of nodes & 10 & 30 \\
Positive learning rate & 0.4 & 0.5 \\
Negative learning rate & 0.01 & 0.05 \\
Weights learning rate & 0.15 & 0.2 \\
Learning Decay & 0.000001 & 0.00002 \\
Number of epochs & 5000 & 10000 \\
\hline
\end{tabular}
\end{table}
\begin{table}[ht!]
\small
\renewcommand{\arraystretch}{1.3}
\caption{Parameter Ranges for SVM}
\label{tab:svm-params}
\centering
\begin{threeparttable}
\begin{tabular}{lcc}
\hline
\bfseries  Parameters & \bfseries min & \bfseries max\\
\hline
C & 0.1 & 10 \\
Kernel Function$^1$ & 1 & 4 \\
Degree of polynomial kernel function & 3 & 5 \\
Gamma of kernel functions 2, 3 and 4 & 0.1 & 1 \\
Independent term in kernel functions 2 and 3 & 0.01 & 1 \\
\hline
\end{tabular}
\begin{tablenotes}
    	\item\footnotesize{$^1$1: linear, 2: poly, 3: rbf and 4: sigmoid.}
    \end{tablenotes}
\end{threeparttable}
\end{table}
\begin{table}[ht!]
\small
\renewcommand{\arraystretch}{1.3}
\caption{Parameter Ranges for MLP}
\label{tab:mlp-params}
\centering
\begin{threeparttable}
\begin{tabular}{lcc}
\hline
\bfseries  Parameters & \bfseries min & \bfseries max\\
\hline
Number of neurons in each layer & 1 & 100 \\
Number of hidden layers & 1 & 3 \\
Learning rate & 0.001 & 0.1 \\
Momentum & 0.85 & 0.95 \\
Epochs & 100 & 200 \\
Optimizer$^1$ & 1 & 3 \\
Activation function$^2$ & 1 & 3 \\
Learning Decay$^3$ & 1 & 3 \\
\hline
\end{tabular}
    \begin{tablenotes}
    	\item\footnotesize{$^1$1: lbfgs; 2: sgd; 3: adam}; \footnotesize{$^2$1: logistic; 2: tanh; 3: relu}; \\ \footnotesize{$^3$1: constant; 2: invscaling; 3: adaptative}.
    \end{tablenotes}
\end{threeparttable}
\end{table}
\begin{table}[ht!]
\small
\centering
\begin{threeparttable}
\renewcommand{\arraystretch}{1.3}
\caption{Parameter Ranges for Label Spreading and Label Propagation}
\label{tab:spread-prop-params}
\centering
\begin{tabular}{lcc}
\hline
\bfseries  Parameters & \bfseries min & \bfseries max\\
\hline
Kernel Function$^1$ & 1 & 2 \\
$\gamma$ (for RBF Kernel) & 10 & 30 \\
Number of Neighbors (for KNN Kernel) & 1 & 100 \\
$\alpha^*$ & 0 & 1 \\
Number of epochs & 20 & 100 \\
\hline
\end{tabular}
\begin{tablenotes}
\footnotesize\item\footnotesize$^1$1: RBF and 2: KNN. * $\alpha$ is only used for label spreading.
\end{tablenotes}
\end{threeparttable}
\end{table}

We considered a projected clustering problem, where each sample should be assigned to a single cluster, and SS-SOM was set to operate in such mode. For classification purposes, if available, we use the node class as the predicted class. Otherwise, it is straightforwardly considered as an error. The next section presents the obtained results and their analysis.

\begin{table}[ht!]
\centering
\begin{threeparttable}
\renewcommand{\arraystretch}{1.3}
\caption{Parameter Ranges for SS-SOM}
\label{tab:sshsom-params}
\centering
\begin{tabular}{lcc}
\hline
\bfseries  Parameters & \bfseries min & \bfseries max\\
\hline
Activation threshold ($a_t$) & 0.80 & 0.999 \\
Lowest cluster percentage (lp) & 0.001 & 0.01 \\
Relevance rate ($\beta$) & 0.001 & 0.5 \\
Max competitions ($age\_wins$) & $1 \times S^*$ & $100 \times S^*$ \\
Winner learning rate ($e_b$) & 0.001 & 0.2 \\
Wrong winner learning rate ($e_w$) & $0.01 \times e_b$ & $1 \times e_b$ \\
Neighbors learning rate ($e_n$) & $0.002 \times e_b$ & $ 1 \times e_b$ \\
Relevance smoothness ($\epsilon \beta$) & 0.01 & 0.1 \\
Connection threshold ($minwd$) & 0 & 0.5 \\
Number of epochs ($epochs$) & 1 & 100 \\
\hline
\end{tabular}
\begin{tablenotes}
\small\item * \textit{S} is the number of input patterns in the dataset.
\end{tablenotes}
\end{threeparttable}
\end{table}

\subsection{Experimental Results and Analysis}
\label{exp-results}

\begin{figure*}[ht!]
  \centering 
  \subfigure[]{\includegraphics[width=0.32\linewidth,scale=1]{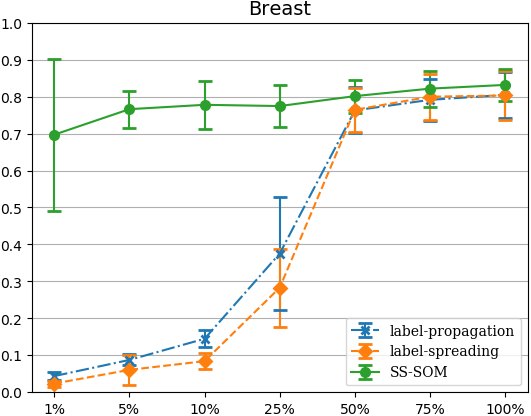} 
  \label{fig:breast}} 
  \subfigure[]{\includegraphics[width=0.32\linewidth,scale=1]{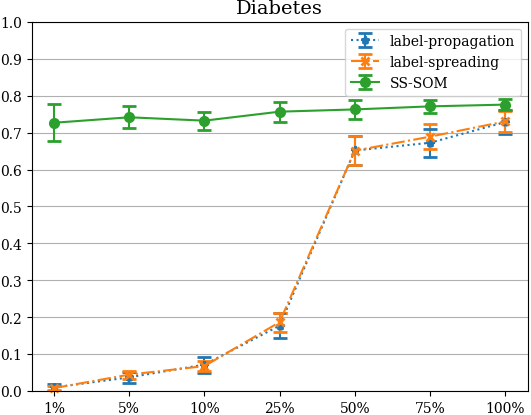} 
  \label{fig:diabetes}} 
  \subfigure[]{\includegraphics[width=0.32\linewidth,scale=1]{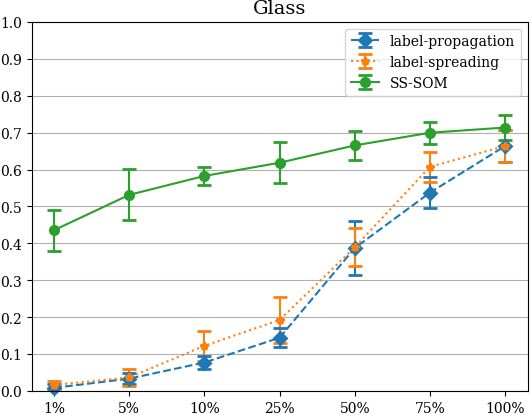}
  \label{fig:glass}} 
  
  \centering 
  \subfigure[]{\includegraphics[width=0.32\linewidth,scale=1]{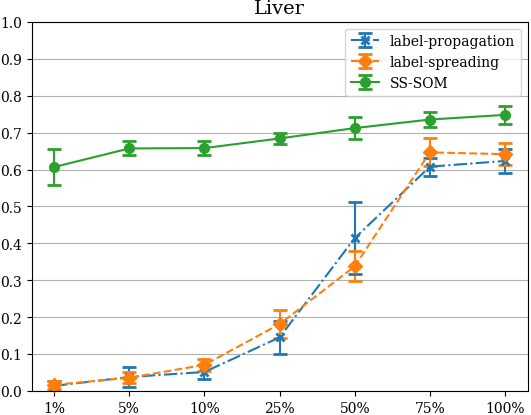}
  \label{fig:liver}} 
  \subfigure[]{\includegraphics[width=0.32\linewidth,scale=1]{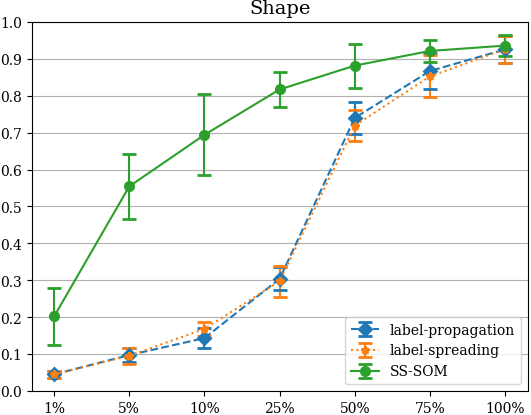}
  \label{fig:shape}} 
  \subfigure[]{\includegraphics[width=0.32\linewidth,scale=1]{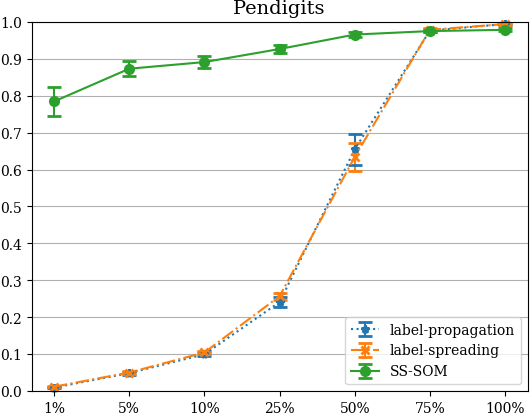}
  \label{fig:pendigits}} 
  \caption{Best mean accuracy and standard deviation as function of the percentage of supervision for each dataset}	
  \label{fig:results}
\end{figure*}

\fref{fig:results} shows the results of SS-SOM in comparison with Label Propagation and Spreading for the real-world datasets as a function of the percentage of labeled data. In all datasets, the performance of the proposed method is superior to the other semi-supervised methods concerning the supervision rate between 1\% to 75\%, whereas with higher percentages (100\%) the difference is smaller, but it continues to outperform or obtain comparable results. These results show the robustness of proposed method in situations when only a small number of labeled data is available.

\tref{tab:results} shows the results of SS-SOM and other semi-supervised methods using 100\% of the labeled data, allowing a fair comparison with supervised methods such as GRLVQ, MLP, and SVM. Our method shows a comparable performance with the other semi-supervised methods, where the biggest difference is for Vowel. Also, SS-SOM appears as the best overall among the semi-supervised methods (the first three in the table), as well as the MLP among the supervised (the last three in the table).

On considering all methods at 100\% of supervision, the MLP outperforms all the others in four of seven datasets. Our method presented the best result for the Shape dataset, outperforming all the other methods. Whereas Label Spreading and Propagation methods are the best ones for Vowel, the SS-SOM showed better results than two of three supervised methods, SVM and GRLVQ, with the former showing a low accuracy value. Also, the SVM appears as the best for Pendigits. Besides that, in all the other datasets, SS-SOM showed results close to the best, even with it not being the primary objective of this work.

\begin{table*}[ht]
\small
\centering
\begin{threeparttable}
\renewcommand{\arraystretch}{1.3}
\caption{Accuracy Results for Real-World Datasets with 100\% of the labeled data}
\label{tab:results}
\centering
\begin{tabular}{c||ccccccc}
\hline
\bfseries  Accuracy & \bfseries Breast & \bfseries Diabetes & \bfseries Glass & \bfseries Liver & \bfseries Pendigits & \bfseries Shape & \bfseries Vowel\\
\hline\hline
SS-SOM & \textbf{0.832 (0.044)} & \textbf{0.776 (0.016)} & \textbf{0.714 (0.033)} & \textbf{0.748 (0.025)} & 0.978 (0.004) & \underline{\textbf{0.935 (0.029)}} & 0.876 (0.017) \\
Label Propagation & 0.805 (0.063) & 0.730 (0.031) & 0.663 (0.044) & 0.623 (0.036) & \textbf{0.994 (0.003)} & 0.925 (0.036) & \underline{\textbf{0.948 (0.012)}}\\
Label Spreading & 0.805 (0.066) & 0.729 (0.031) & 0.663 (0.044) & 0.640 (0.031) & \textbf{0.994 (0.003)} & 0.925 (0.036) & \underline{\textbf{0.948 (0.012)}}\\
\hline
MLP & \underline{\textbf{0.854 (0.032)}} & \underline{\textbf{0.791 (0.017)}} & \underline{\textbf{0.746 (0.031)}} & \underline{\textbf{0.766 (0.031)}} & 0.993 (0.001) & 0.923 (0.034) & 0.874 (0.033) \\
SVM & 0.850 (0.037) & 0.788 (0.020) & 0.718 (0.028) & 0.746 (0.054) & \underline{\textbf{0.997 (0.001)}} & \textbf{0.931 (0.030)} & \textbf{0.909 (0.022)} \\
GRLVQ & 0.830 (0.049) & 0.772 (0.020) & 0.676 (0.027) & 0.699 (0.022) & 0.915 (0.004) & 0.823 (0.061) & 0.515 (0.027) \\ 						
\hline
\end{tabular}
\begin{tablenotes}
\centering
\footnotesize\item\footnotesize In bold, the best results for each dataset on each category: semi-supervised and supervised methods. The underlined results indicate the global best.
\end{tablenotes}
\end{threeparttable}
\end{table*}

\section{Conclusion and Future Work}
\label{conclusions}
This article presented an approach for classification and clustering with semi-supervised learning. The behavior of SS-SOM was shown to have led to significant improvements in classification results for small amounts of labeled data, establishing its position as a good option when dealing with such problems, which is the central point of this article. The proposed method showed its robustness under this condition, being better than other semi-supervised models, achieving impressive results even with only 1\% of labeled data. Furthermore, despite the fact that SS-OM has 11 parameters, only three of them ($a_t$, $lp$ and $e_b$) present important effects on the results.

Also, in all datasets, using 100\% of the labels, SS-SOM showed results better than or at least close to the best found in comparison with others supervised and semi-supervised methods, even with it not being the objective of this work. 

It is important to mention that in the current implementation, the self-organizing process is run for a number of epochs sampled from LHS, which is usually greater than the necessary to converge, even at the defined interval. An adequate stop criterion is an object of study for future versions in order to reduce the training time.

Notice that LARFDSSOM presented good results for subspace clustering \cite{bassani2015larfdssom}, and when there is no labeled sample available, SS-SOM works exactly as LARFDSSOM, inheriting its characteristics and performance. However, when labeled samples are given, the results can be even better. Moreover, with a small change, SS-SOM could also incorporate reinforcement learning, being, thus, capable of switching between three different learning approaches, to exploits several forms of information available, what is left for future work.

\section*{ACKOWLEDGMENTS}
The authors would like to thank the Brazilian National Council for Technological and Scientific Development (CNPq) for supporting this research study.


\bibliographystyle{IEEEtran}
\bibliography{references}

\end{document}